\documentclass[letterpaper, 10 pt, conference]{ieeeconf}  

\IEEEoverridecommandlockouts                              

\usepackage{graphics} 
\usepackage{times} 
\usepackage{amsmath} 
\usepackage{amssymb}  
\usepackage{color}
\usepackage{booktabs}
\usepackage{soul}
\usepackage{algorithm}
\usepackage{algorithmic}

\usepackage{makecell}
\usepackage{tabularx}

\usepackage{color} 
\usepackage{url} 
\usepackage{booktabs}
\usepackage{listings}
\usepackage{pbox}

\usepackage{pgfplots}
\usepackage{subfig}

\newcommand{\etal}{{\em et al.}}
\newcommand{\at}{\makeatletter @\makeatother}

\title{\LARGE \bf
Modality-Buffet for Real-Time Object Detection
}

\author{Nicolai Dorka*$^{1}$, Johannes Meyer*$^{1}$, and Wolfram Burgard$^{1,2}$
\thanks{This work has been supported by the Freiburg Graduate School of Robotics.}
\thanks{*Contributed equally}
\thanks{$^{1}$All authors are with the Department of Computer Science, University of Freiburg, Germany.
        {\tt\small {dorka,meyerjo,burgard}@cs.uni-freiburg.de}}%
\thanks{$^{2}$Wolfram Burgard is also with the Toyota Research Institute, Los Altos, USA.}%
}

\begin{document}

\maketitle
\thispagestyle{empty}
\pagestyle{empty}

\begin{abstract}
Real-time object detection in videos using lightweight hardware is a crucial component of many robotic tasks.
Detectors using different modalities and with varying computational complexities offer different trade-offs.
One option is to have a very lightweight model that can predict from all modalities at once for each frame.
However, in some situations (e.g., in static scenes) it might be better to have a more complex but more accurate model and to extrapolate from previous predictions for the frames coming in at processing time.
We formulate this task as a sequential decision making problem and use reinforcement learning (RL) to generate a policy that decides from the RGB input which detector out of a portfolio of different object detectors to take for the next prediction.
The objective of the RL agent is to maximize the accuracy of the predictions per image.
We evaluate the approach on the Waymo Open Dataset and show that it exceeds the performance of each single detector.
\end{abstract}



\section{Introduction}
\label{sec:introduction}

Object detection is a key component of nowadays robotic platforms as they get more and more integrated into the real world and have to be able to react to dynamic objects as humans, cyclists or cars. To carry out a reaction to a dynamic object the robot has to first of all perceive it. Different approaches have been developed to tackle the problem of object detection~\cite{ren2017faster}\cite{he2017mask}\cite{redmon2018yolov3}. The proposed solutions vary in their requirements regarding resources and run-time as well as in their performance.

Results obtained by state-of-the-art object detectors are often remarkable but rely on high-end computers. Using such computers on mobile robots is not realistic as these robots are often powered via battery or cannot carry such huge payload. For that reason, products such as the NVIDIA Jetson AGX Xavier \cite{nvidia_xavier},  which offer state-of-the-art GPU techniques at a very small scale and power consumption, have been developed. However, when running a Faster R-CNN with a ResNet 50-FPN on an NVIDIA Jetson AGX Xavier, it takes $340$ ms to get the bounding box detections. In comparison the same forward pass on an NVIDIA Titan X that is commonly used in high-end computers takes $80$ ms. This restricts the ability of robotic platforms to use the network at high-frame rates which is required for interaction with dynamic objects.

In order to assure robustness of the method against failure under varying conditions it is desirable that the detector uses different sensor modalities.
However, in deep networks the encoding and fusion of additional modalities in a way that substantially increases the performance over a model that uses only one modality further increases its computational cost considerably \cite{pfeuffer2018optimal}. 
At the same time in many scenes one modality may be sufficient for a good detection of objects - for example RGB for images in clear daylight and lidar during the night - as the benefits of the modality in this scene-type outweigh the disadvantages of another modality.

\begin{figure}[t]
    \centering
    \includegraphics[width=\linewidth]{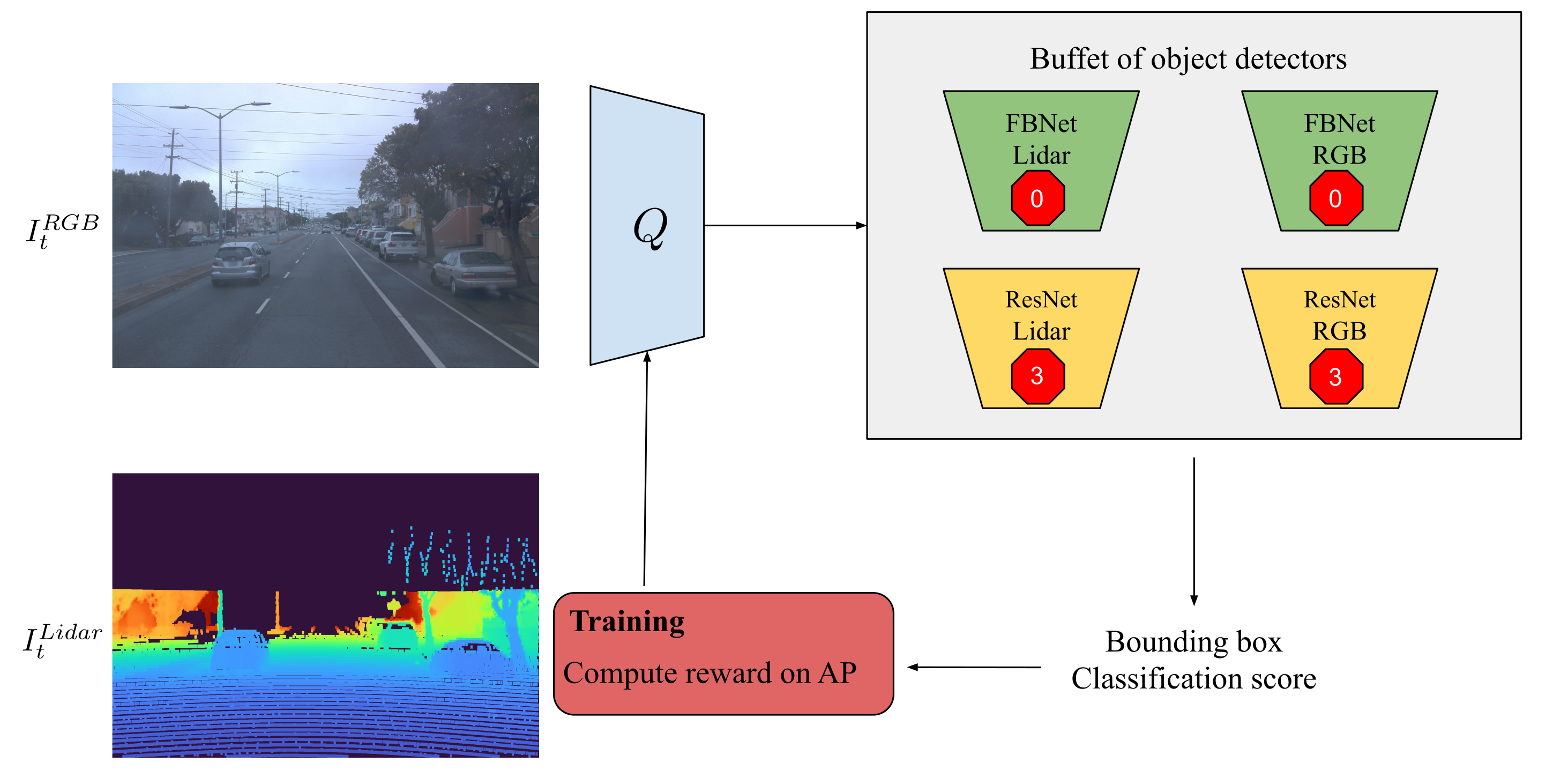}
    \caption{
        Our proposed Reinforcement Learning (RL) agent is at each timestep $t$ faced with the question which of the modalities and available object detectors to choose from a buffet of options. Each modality and detector has a different strength or weakness and future frames are impacted by the decision of the agent at time $t$. For example, choosing a ResNet based feature detector blocks the agent for the next three frames as the embedded hardware does not allow for a fast computation of this detector. The accuracy of the resulting detections is used as a reward signal for the reinforcement learning algorithm in order to optimize the policy.
    }
    \label{fig:coverfigure_approach}
\end{figure}

In this work we propose a new lightweight method which uses the image recorded by an RGB camera to choose from a portfolio of different detectors that vary in computational complexity and in the kind of sensor modalities they use.
Each detector offers some scenarios where it is suited best given its unique trade-off between computational cost and accuracy.
As more complex detectors require more time to generate predictions, not every frame can be processed in real-time if they are used.
The best that can be done for the frames coming in until the detector outputs its prediction is to extrapolate from the most recent previous prediction. 
Still this might sometimes be the preferable choice if the extrapolation is easy (e.g., in static scenes) as the prediction of larger models are usually more accurate.
Choosing from detectors with different prediction generation times is a sequential decision making process where the current action influences the next state at which a decision has to be made.
We use reinforcement learning (RL) to learn a policy that decides about the size of the model for the detector and about which modality should be used in order to maximize the overall accuracy of the predictions for all incoming frames.

The approach is evaluated on data provided in the Waymo Open Dataset (v1.0) \cite{sun2019scalability} which contains sequences of images recorded at a frame rate of $10$ Hz.
We demonstrate that it is possible to learn a policy that chooses models in such a way that the accuracy per image exceeds the performance of all the single detectors.

\section{Related Work}

\subsection{Object detection}

\subsubsection{Real-time}

Recently, since the introduction of YOLO and its most recent successors~\cite{redmon2018yolov3} object detection using just a single-stage instead of multiple processing stages is an active research area. These object detectors promise a real-time execution combined with a good performance. 
Even more recently, these one-stage object detectors have been improved by more elaborate strategies to fuse features in a feature pyramid network \cite{tan2019efficientdet}. Zhao \etal \cite{zhao2019m2det} propose another method to extract more suitable features from a pyramid network. Also novel concepts to tackle the problem of single-stage object detection have been proposed~\cite{zhou2019objects}\cite{law2018cornernet}.

However, these approaches are still designed for the computation on desktop GPUs and are thus not suitable for the computation on embedded hardware. Therefore, we use an efficient feature extractor that was specifically designed for the usage in embedded hardware. As trade-off between the encoder speed and the prediction accuracy we use a Faster R-CNN implementation as detection algorithm. 

\subsubsection{Videos}
Detecting objects from a video is an interesting topic, as in contrast to the standard object detection task the temporal information is added to the input. Several different approaches have been proposed in the literature.

Kang \etal \cite{kang2018t} make use of the temporal information during the training of the object detectors. Predicted bounding boxes are propagated to the next frame and then used for refinement and supplemented by new detections. However, they use a very deep feature extractor that makes the method not applicable in real-time. Other temporal linking techniques have been proposed by Tang \etal \cite{tang2019object}.

A different strategy for the problem of object detection in videos has been proposed by Zhu \etal \cite{zhu2017flow}, who use the optical flow in consecutive frames in order to get a detection at frame $t$. However, their method requires the information from future frames and thus is also not real-time capable.

\subsubsection{Different modalities}
Having a multitude of input modalities at hand, different ways on how to fuse the distinct information have been studied and analyzed. Dietmayer \etal \cite{pfeuffer2018optimal} investigate at which stage of the detection process the best results through fusion can be obtained.
Du \etal \cite{du2018general} use the RGB image to find objects within the 2d image and then use these bounding boxes to extract data from the respective position in the Lidar scan. The extracted Lidar data is used to create a final 3d bounding box detection. While focusing on the detection of 3d bounding boxes, this approach is the current state-of-the-art on the KITTI car detection benchmark \cite{Geiger2012CVPR}. However, the inference time is rather slow according to the official benchmark. A similar concept of combining RGB and Lidar data has been proposed by Qi \etal \cite{Qi_2018_CVPR}.
Mees \etal \cite{Mees2016iros} proposed to learn to weight class embeddings from different modalities to improve object detection.

\subsection{Reinforcement learning for speed accuracy tradeoff}

Using RL for the trainng of a policy to trade-off speed and accuracy has been proposed for different tasks. For object tracking it was proposed to learn a policy to select each frame between models of different computational cost \cite{huang2017learning}.

Chinchali \etal~\cite{chinchali2019network} use a RL algorithm in order to learn whether a robot should ``off-load'' a classification problem to a remote machine. At each time step the RL agent has to decide whether to query the model over the network, keep the old prediction, or use its own prediction module. The reward signal then consists of the model error and the latency/compute costs of the detection.

Other works learned for the task of semantic segmentation if to use a model for a frame or interpolate from the temporally surrounding predictions under computation constraints \cite{Mahasseni2017BudgetAwareDS}. Interpolation is only applicable in hindsight and not in real-time applications as future frames are not available at a given time.

For object detection it was proposed to learn a model that decides on every frame if a detection or tracking model should be used \cite{luo2019detect}. While the classification model used in that work is interpreted as an RL model this is only meaningful in so far as any classification method can be framed as an RL algorithm where the action in a state does not influence the next state.

Most closely related to our work is that of Liu \etal \cite{2019arXiv190310172L}
who use a memory that is a version of an LSTM \cite{lstm} for object detection from RGB images in videos.
Each frame a policy decides if for the next frame either a slow or a fast feature embedding network is selected. The LSTM takes the embedding and its internal state to produce a refined embedding which is used for prediction. Only the slow network updates the internal state of the LSTM. The policy is learned with a reward signal that incorporates a speed and a accuracy reward.
Except from the fact that they considered only one modality the main differences to our method are that we define the reward signal only via the accuracy and that we decide for the current and not the next frame which model to use.

\section{Method}
\label{sec:method}

Our method consists of two parts. First, a set of object detectors that have different computational complexities and that take different types of sensor modalities as input. Second, a reinforcement learning algorithm that takes as input an RGB image and selects one of the detectors. The overall approach is displayed in Figure \ref{fig:coverfigure_approach}.
We demonstrate our approach for the setting in which there are two modalities available, an RGB camera image and an image from projected lidar data. Furthermore, we use object detectors of two different sizes.
However, it should be noted that our approach is not specific to this particular setting and is also applicable in settings with other sets of detectors which could use more or other modalities.

\subsection{Object Detection}
We use object detectors of two different sizes. The first class uses as a network architecture a ResNet-50 with a feature pyramid network \cite{he2016deep}\cite{lin2017feature} and a Faster R-CNN implementation to detect objects. 
The second class is the same as the first one except that it uses instead of the ResNet the FBnet by Wu \textit{et al.} \cite{wu2019fbnet}, which is more lightweight and specifically designed to operate on mobile low-energy consuming devices.
For each of the two modalities RGB and lidar we train a model from each network class using the respective modality as input
resulting in a total of four different object detectors.

While in this work Faster R-CNN models are used, the same method could also be applied on top of differently sized one-stage methods or even on top of a combination of one-stage and multi-stage models.

\begin{figure}
    \centering
    \includegraphics[width=\linewidth]{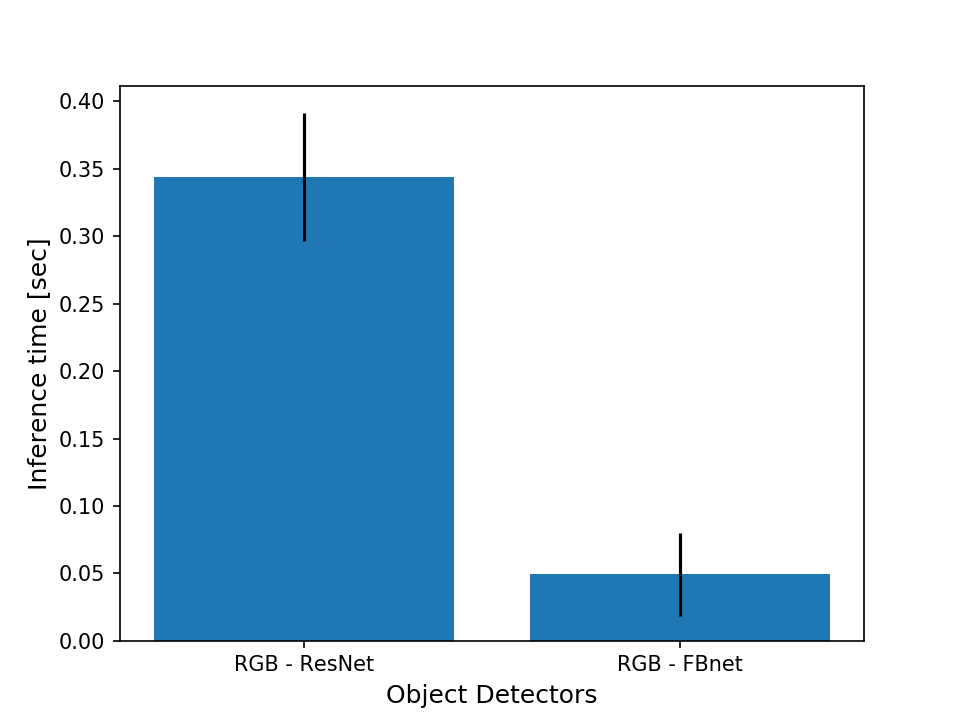}
    \caption{Different network architectures have different inference times on an NVIDIA Jetson Xavier. The x-axis shows the different models trained on the RGB modality, the y-axis shows the inference time in seconds. The black line on top of each bar indicates the standard deviation in between runs. The average run-time is calculated over all images on the used test-set. The timing data is almost identical for the Lidar images.
    }
    \label{fig:inference_time}
\end{figure}

In Figure \ref{fig:inference_time} we show the inference times for the two different model classes on the NVIDIA Jetson Xavier. The FBnet based model can reliably predict every frame when they are perceived at $10$Hz even if the variance in inference time is incorporated. If the much larger ResNet-50 is used, a prediction can only obtained reliably for every fourth frame at an equal frame rate.

At the same time the predictions of the ResNet based model are more accurate and in the next section we describe how we frame the problem of choosing the right model at the right moment in a reinforcement learning setting.

\subsection{Reinforcement Learning}

In RL an agent interacts with its environment for a discrete set of timesteps $T$. During each step $t \in T$ the agent is in a state $s_t \in S$ and has to select an action $a_t \in A$. The agent receives a scalar reward $r_t$ for the action and transitions to the next state $s_{t+1}$ that depends on the chosen action.
The goal is to learn a policy $\pi: S \rightarrow A$ that chooses actions such that the return which is the sum of all received rewards $\sum_{t \in T} r_t$ is maximized.

For the RL algorithm we use a version of Rainbow \cite{Hessel2017Rainbow} which adds several improvements on top of the DQN algorithm \cite{mnih2015humanlevel}. Different from the original Rainbow we do not use prioritized experience replay and for exploration we use $\epsilon$-greedy instead of noisy networks. 
The algorithm learns a state-value function - called the Q-function - which is modeled with a neural network to map the state to the parameters of a distribution of future rewards for each action in that state. The agent chooses the action with the highest predicted expectation over the future returns.
Experiences of the agent are stored in a replay buffer and the learning of the value function happens on samples drawn randomly from that buffer.
For more detailed information about Rainbow we refer to the original publication \cite{Hessel2017Rainbow}.

\begin{figure*}
    \centering
    \includegraphics[width=\textwidth]{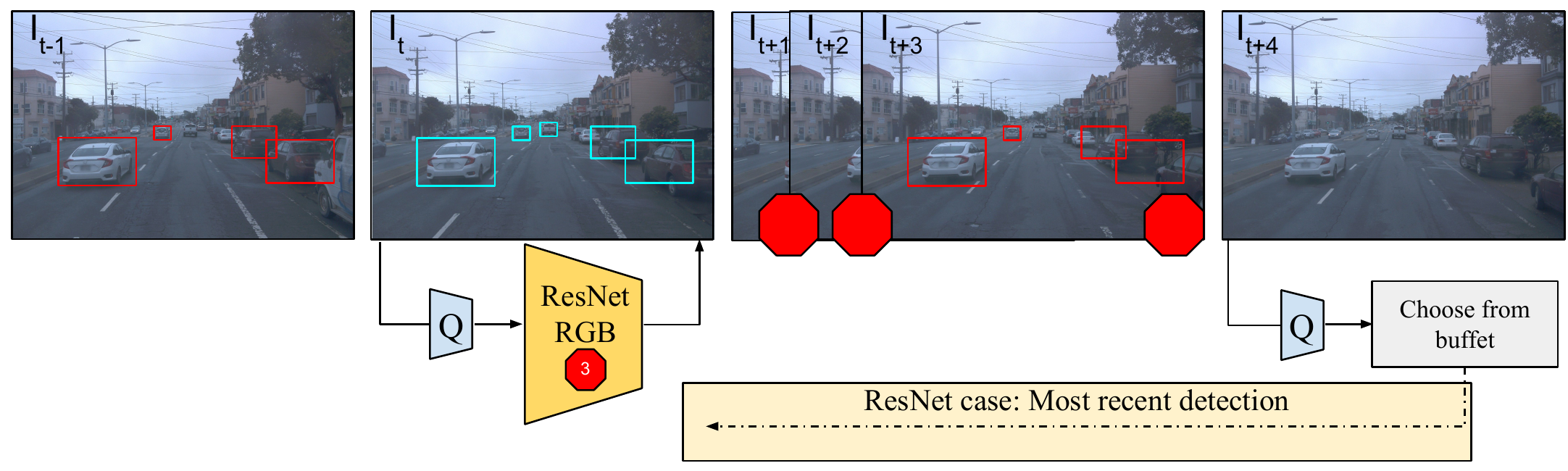}
    \caption{In $I_{t-1}$ the most recent previous detection for this frame are shown. After processing
    $I_t$ the RL agent queries the Resnet-50 and is blocked for the three consecutive frames. Thus,
    it has to rely on the detections from previous frames. At time-step $I_{t+4}$ the ResNet has provided the prediction for the frame it was queried on. This becomes the most recent prediction, if the network decides to query the ResNet in $I_{t+4}$.
    }
    \label{fig:consecutive_frames}
\end{figure*}

We define the environment for the RL agent on top of a dataset that contains sequences of images.
Each sequence is treated as an episode. The state $s_t$ of the agent is the RGB image of the current frame $f_{t'}$ in the sequence downsampled to the spatial dimensions $(84, 84)$.
In each state the agent has to choose the object detector that generates the next prediction.
Depending on the computational requirements of the chosen detector it takes $k \in \mathbb{N}$ frames to get the prediction.
The next state $s_{t+1}$ of the agent is then defined to be the RGB image of the frame $f_{t'+k +1}$  which is $k+1$ steps into the future from the state $s_t$ where the detector was chosen.
This is also visualized in Figure \ref{fig:consecutive_frames}.
The reward signal for taking a specific action $a_t$ in state $s_t$ is defined via the average precision (AP) of the predictions from the chosen detector for the current frame and the extrapolation from the most recent previous prediction for all $k$ frames that come in during the inference time of the model.
Put more concretely, the AP for frame $f_{t'}$ is defined to be the AP of the model prediction for $f_{t'}$.
For the frames $f_{t'+1}, \dots, f_{t'+k}$ the AP is computed from predictions that are extrapolated from the most recent previous prediction as the computation of the model takes $k$ frames.
We define the predictions for these frames to be equal to this previous prediction. 
Note that for the extrapolation more sophisticated methods like tracking of objects could be used, but this is orthogonal to our method and so we concentrated on the simple case.
The reward for the agent is then defined as the sum of the $k+1$ AP values obtained with this procedure.
In more formal terms, the reward for choosing in state $s_t$ the action $a_t$ that corresponds to a model that has an inference time of $k$ frames is defined as
\begin{align}
    r(s_t, a_t) = \text{AP}(f_{t'}, a_t) + 
    \sum_{i=1}^k
    \text{AP}(f_{t'+i}, a_{t-1}),
\end{align}
where $\text{AP}(f, a)$ denotes the AP score between the ground truth for frame $f$ and the predictions of the model corresponding to action $a$ for $f$.

Different from previous work \cite{huang2017learning}\cite{chinchali2019network} our reward signal is defined solely via the accuracy and is not a combination of a speed and a accuracy reward. The tradeoff between these two is incorporated implicitly in the design of the environment setup which is defined such that real-time performance is always guaranteed. 

As the RL agent is trained entirely on the training set of the dataset there is one complicating factor regarding how to generate the model predictions. The object detectors generalization abilities on images not seen during training differ depending on model capacity and other factors. The reward signal for the RL agent would be biased in a misleading way if for the generation of the rewards the predictions from models trained on that example would be taken. The reason is that for example methods that overfit on the training data would be preferred to be chosen by the agent.
To circumvent this problem we split the training data into five different folds and train the networks in a leave-one-out manner. For each image in the training set we have a prediction from an object detector which was not trained on this image. This allows us to use the same dataset for the training of the RL agent as the generalization error of the individual networks is expected to be the same on the individual folds as on the training set. 
In Table \ref{tab:detection_results_individual} we report the average performance of the hold-out trained detectors on the test set and on the hold-out fold. We can see that the trained detectors perform similar on the test set and on their holdout set. Also the strength of the object detector relative to each other stays similar. Moreover, this also holds when the models are trained on the entire training set and evaluated on the test set. From this experiment, we conclude that we can use the object detectors trained on the hold-out set to mimic the detection signal the RL agent will observe on the test set. Thus, we use the output of these detectors to train the RL agent on the training set.
To speed up the training of the RL agent we stored the predictions of the different detectors in a lookup table. That makes the environment for training the RL agent very fast as the model predictions can be simply read out and only the AP scores have to be computed.
To generate the predictions for the test set we trained the object detectors on the whole training set.

During runtime the RL agent requires only a single forward path to choose which detector to take. Since the network of the agent we use is very lightweight with $3$ convolutional layers and one hidden fully-connected layer
the additional computational cost of our method is negligible in comparison to the much larger networks of the detectors.

\section{Experiments}
\label{sec:experiments}

\subsection{Detection on the Waymo Dataset}

\begin{table}[b]
    \centering
    \resizebox{\linewidth}{!}{%
    \begin{tabular}{|c|c|c||c|c|}
        \hline
         Backbone & Modality    & AP Test & \makecell{avg. AP Folds\\ \at Test} & \makecell{avg. AP Folds \\ \at Hold-Out}  \\
         \hline
         ResNet-50 FPN & RGB    & $0.398$ & $0.384$ & $0.411$  \\ 
         FBnet & RGB            & $0.206$ & $0.188$ & $0.194$  \\ 
         \hline
         ResNet-50 FPN & Lidar  & $0.253$ & $0.246$ & $0.278$  \\ 
         FBnet & Lidar          & $0.167$ & $0.151$ & $0.174$ \\ 
         \hline
    \end{tabular}
    }
    \caption{
     Different object detectors trained with different feature extractor architectures and data from different modalities. Each detector is trained on the full training set and evaluated on the test set. The given values are for the AP averaged over the IoU levels from $0.5$ to $0.95$ with a step size of $0.05$. Additionally, the average performance of the individual detectors trained on the hold-out sets is shown.}
     \label{tab:detection_results_individual}
\end{table}

We train different object detectors using the Mask R-CNN Benchmark implementation \cite{massa2018mrcnn} on the Waymo open dataset. We use the initial version of the Waymo Open Dataset, namely v1.0. It consists of $100$ sequences with annotated 2d bounding boxes and $25$ of those annotated sequences are available as test sequences. Data from five cameras are available. In this work, we only use the one of the frontal camera. Each image is also accompanied with the most recent lidar scan. 
For both modalities the RGB image and the lidar scan which we project on the image plane, we train a Faster R-CNN implementation with a ResNet-50 using a feature pyramid, and a FBnet as the respective feature extractor.

In Table \ref{tab:detection_results_individual} the results for the individual object detectors with the different modalities are shown. As expected the different network architectures and the individual modalities have different levels of performance. For all trained object detectors we report the AP levels across different Intersection over Union (\textit{IoU}) thresholds from $0.5$ to $0.95$ with a stepsize of $0.05$. Intersection over Union is a normalized measure which computes the ratio between the area in which two bounding boxes overlap and the total area of both boxes aggregated. The detector based on the Resnet-50 which uses the RGB modality yields an AP of $0.398\%$. This is the best performance of all networks which are trained on the whole training set. The second strongest object detector is the ResNet which was trained on the lidar images. Both networks trained with the FBnet as feature extractor are well behind the ResNet detector.

\subsection{Reinforcement Learning}

\begin{table}[t]
    \centering
    \begin{tabular}{|l|c|c|c|}
        \hline
        Model & Frames \\
        \hline
        FBnet Lidar & $0$ \\
        FBnet RGB &  $0$ \\
        ResNet50 FPN Lidar & $3$ \\
        ResNet50 FPN RGB & $3$ \\
        \hline
    \end{tabular}
    \caption{Frame-timout for the agent after a certain model is chosen.}
    \label{tab:skipped_frames}
\end{table}

The RL agent was trained for $300,000$ environment steps where the exploration parameter $\epsilon$ was linearly decayed from $1.0$ to $0.01$ over the $300,000$ steps. The target network was updated every $8,000$ steps and the training started after the replay buffer was filled with $20,000$ transitions. All other hyperparameters and the network architecture are the same as in the original Rainbow paper.
We considered three different kinds of IoU thresholds ($0.5, 0.7, 0.5:0.95$) for the computation of the AP in the reward signal and trained one agent for each of these.

We report for each model the number of frames for which the most recent prediction has to be used because of the models inference time in Table \ref{tab:skipped_frames}. When using one of the FBnet models the agent can operate again in the next frame while for the models using the ResNet50 $3$ frames have to be bridged with the previous prediction.

\subsection{Baselines}
We compare our method to multiple types of baselines.
The first set of baselines consists of policies that always choose exclusively one of the four single models. In the following they are named after the respective model they use.
Note that the strategies using the ResNet based models can not produce a prediction for every frame and rely on the previous prediction in the meantime which reduces their performance compared to the case where the models predictions are computed for every frame.
A second set of baselines uses a random policy to choose the model to execute. We also evaluated a variant of this policy as we exclude the \textit{ResNet50-Lidar} from the set of modalities as this model gives with the long prediction time substantially worse results than the other methods. 
Furthermore, one strategy that sequentially picks all the models and a variant of it without the \textit{ResNet50-Lidar} model is evaluated.
Lastly, we evaluate against a heuristic which uses the average color of the image to decide which object detector to use.
The idea is to choose an RGB based model during daytime and a lidar based model when it is dark.
The heuristic selects the \textit{FBnet-Lidar} if the mean pixel value is below a certain threshold and otherwise the \textit{ResNet50-RGB} is used.
For each of the three reported AP metrics we evaluated the heuristic on the test set for $10$ evenly spaced pixel value thresholds between $0$ and $255$ and report the best performance. 
As the threshold hyperparameter is fitted on the test set this heuristic can be seen as an oracle about how much performance can be gained if only the lighting is taken into account to choose a model.

\subsection{Evaluation Metric}
Metrics which evaluate every frame such as PASCAL VOC \cite{PASCALVOC} or Microsoft COCO \cite{MSCOCO} put implictly more weight on frames with a higher amount of bounding boxes. This benefits the deeper, stronger object detectors as they are able to recall more of these boxes with a higher precision. However, in the application studied in this work, specifically detecting objects in a sequence of constantly appearing frames it is more important to have an on average higher AP per frame than an higher AP over all detections in the dataset. For that reason, we compute the AP score per image and compute the mean over these scores as our evaluation metric for the following experiments. For the few images without ground truth bounding boxes  ($3\%$ in the test set) we return $1$ if no bounding box is predicted and $0$ otherwise.

\subsection{Results}

\begin{table}[t]
    \centering
    \begin{tabular}{|l|c|c|c|}
        \hline
        Model                & AP\at0.7 & AP\at0.5 & AP\at0.5:0.95\\
        \hline
        FBnet Lidar          & $0.213$ & $0.285$ & $0.179$ \\
        FBnet RGB            & $0.220$ & $0.333$ & $0.195$ \\
        ResNet50 FPN Lidar   & $0.137$ & $0.212$ & $0.117$ \\
        ResNet50 FPN RGB     & $0.214$ & $0.350$ & $0.196$ \\
        \hline
        Random                          & $0.183$ & $0.281$ & $0.159$ \\
        Random w/o ResNet Lidar         & $0.202$ & $0.317$ & $0.181$ \\
        Alternating                     & $0.171$ & $0.272$ & $0.150$ \\
        Alternating w/o ResNet Lidar    & $0.197$ & $0.308$ & $0.173$ \\
        Lighting heuristic             & $0.223$ & $0.350$ & $0.201$ \\
        \hline
        Learned policy & $\mathbf{0.245}$ & $\mathbf{0.366}$  & $\mathbf{0.214}$ \\
        \hline
    \end{tabular}
    \caption{Results on the test set of the Waymo Open Dataset for different IoU thresholds in the computation of the AP.}
    \label{tab:evaluation_results_method}
\end{table}

We present in Table \ref{tab:evaluation_results_method} the results for all considered methods when evaluated on the test set. 
In the scenario we described in this work, where the GPU is blocked for a certain amount of time when the large model is executed, the best performing individual model  at an IoU level of $0.7$ is the \textit{FBnet RGB}. For the other two IoU levels the \textit{ResNet50 RGB} achievs the highest AP score. 
Both the random  and the alternating selection strategy are not able to outperform one of the individual object detectors they select from. The best of them is the Random agent w/o \textit{ResNet Lidar} with an average AP per frame of $0.202$ at an $0.7$ IoU level.
The only strategy reaching better results than the single models is the lighting heuristic with AP scores of $0.223$, $0.350$ and $0.201$ at the three different IoU levels.
The policy learned using the methodology presented in this work achieves at an IoU level of $0.7$ a mean AP score of $0.245$ on the test dataset outperforming all other considered methods. The same holds for the other two considered IoU thresholds for which the policy reaches AP scores of $0.366$ and $0.214$.

\subsubsection*{Discussion}
From the experiments we can see that the learned policy indeed learned a strong strategy as our method outperforms each of the single models for all of the IoU thresholds and also all the other baselines.
The random and the alternating policy are much worse than the learned policy. However, their performance is degraded from using in $25\%$ of the cases the \textit{ResNet50 Lidar} model which performs poorly. Interestingly, even when this model is removed, the performance of the random and the alternating policy is worse than the worst of the three remaining single models.
Hence, it is not trivial to generate a policy that outperforms each of the single models.
A possible explanation why the random policy is worse than each of the single models might be that when first one of the not so accurate FBnet models is used and afterwards the \textit{ResNet50 RGB}, the most recent previous prediction coming from the FBnet is not so accurate as when it came from the ResNet. 

Furthermore, the results for the lighting heuristic show that some performance can be gained by selecting the lidar or RGB model depending on the lighting. However, this alone can not explain the performance of the learned policy as its improvements over the baselines are much larger indicating that the policy learned a non-trivial strategy.

\subsection{Discussion: Modality usage by policy}

In this section we take a closer look at the three learned policies which got their reward based on three different IoU thresholds. We show the percentage of how often the learned policy uses each of the four models on average for all of the three IoU thresholds in Table \ref{tab:model_percentages}.

\begin{table}[b]
    \centering
    \resizebox{\linewidth}{!}{%
        \begin{tabular}{|l|c|c|c|c||c|}
            \hline
            Policy@IoU   & \makecell{FBnet\\Lidar} & \makecell{FBnet\\RGB}  & \makecell{ResNet50 \\Lidar} & \makecell{ResNet50\\RGB} & AP@IoU\\
            \hline
            0.7 & $42\%$ & $19\%$  & $1\%$  & $38\%$ & $0.245$\\
            0.5 &  $16\%$  & $31\%$ & $0\%$ & $53\%$ & $0.366$\\
            0.5:0.95 & $21\%$ & $35\%$  & $1\%$ & $43\%$ & $0.214$\\
            \hline
        
            \hline
        \end{tabular}
    }
        \caption{Percentage of how often which modality is chosen from the modality-buffet by the learned RL policies. Each of the three policies was trained with the reward signal defined via the AP at a different IoU threshold.}
    \label{tab:model_percentages}
\end{table}

It can be observed that the percentage with which a model is picked does not necessarily align with the performance of the models taken on their own at the given IoU threshold. For example, the results in Table \ref{tab:evaluation_results_method} show that at the IoU threshold of $0.7$ the \textit{FBnet RGB} has the strongest performance when always selected. Despite that fact, the learned policy chooses it less often than the \textit{FBnet Lidar} or the \textit{ResNet RGB}.

\subsection{Ablation Study with two Models}

We conducted a further experiment where we trained a policy to choose only between the \textit{ResNet50 RGB} and the \textit{FBnet Lidar}. The results can be found in Table \ref{tab:evaluation_results_2_models}. 
The learned policy outperforms for all considered IoU thresholds both the single models and the two policies which choose with a random or alternating strategy.
Similarly to above the random strategy is worse than each of the single models which could be explained in the same way as before. However, the learned policy for the two models does not reach the performance of the learned policy for the four models.
This indicates that the proposed method becomes stronger the more models it has in its portfolio.
Thus, incorporating further models, potentially from other modalities, or models that take both RGB and depth as input could be added to likely increase the performance further.

\begin{table}[t]
    \centering
\begin{tabular}{|l|c|c|c|}
    \hline
    Model                & AP@0.7 & AP@0.5 & AP@0.5:0.95\\
    \hline
    FBnet Lidar          & $0.213$ & $0.285$ & $0.179$ \\
    ResNet50 FPN RGB     & $0.214$ & $0.350$ & $0.196$ \\
    \hline
    Random              & $0.206$ & $0.314$ & $0.181$ \\
    Alternating         & $0.192$ & $0.288$ & $0.168$ \\
    \hline
    Learned policy &  $\mathbf{0.242}$ & $\mathbf{0.355}$  & $\mathbf{0.211}$ \\
    \hline
\end{tabular}
    \caption{Results on the testset of the Waymo Open Dataset for different IoU thresholds when only the two models FBnet Lidar and ResNet50 RGB are available.}
    \label{tab:evaluation_results_2_models}
\end{table}
\section{Conclusion}

In this work, we presented a novel RL based method which selects an object detector from a buffet of available detectors in order to maximize the average precision across a sequence of consecutive frames under the requirement of real-time performance. The strong results compared to the single models show that the learned policy is able to both select the right modality and reliably identify those frames in which a more accurate, but slower detection can be queried and in which frames a faster detection is required. 
At the same time we showed that it is not easy to hand-design a policy that outperforms the single models.

Results of experiments with a smaller portfolio of models indicate that the performance of our method likely increases further when the size of the portfolio gets larger and more diverse. This opens the door for many future applications and we believe that our method is especially promising for applications where the conditions vary a lot as the policy selects the model that can cope best with the specific condition.

Another advantage of our method is that if the AP score is not the only important objective further signals could be included in the reward function. One possible objective could be for example the power consumption of the device.

\bibliographystyle{unsrt}
\bibliography{inc/related_work.bib}

\end{document}